%% file: ral2019-yshin.tex
\documentclass[letterpaper, 10 pt, conference]{ieeeconf}  % Comment this line out if you need a4paper

\IEEEoverridecommandlockouts                              % This command is only needed if
\usepackage[pdftex]{graphicx}
\usepackage{amsmath} % assumes amsmath package installed
\usepackage{amssymb}  % assumes amsmath package installed
\usepackage{subfigure}
\usepackage{multirow}
\usepackage{array,booktabs}
\usepackage{diagbox}
\usepackage{balance}
\usepackage{cite}
\usepackage{booktabs}
\usepackage{hyperref}
\usepackage{irap_SIunits}
\usepackage{irap_acronyms}
\usepackage{irap_math}
\usepackage{irap_misc}

\usepackage{soul,color}

\usepackage{xcolor}

\newcommand{\ie}{\textit{i}.\textit{e}.}

\DeclareMathOperator*{\argmin}{argmin}
\DeclareMathOperator*{\argmax}{argmax}

\title{\LARGE \bf
Sparse Depth Enhanced Direct Thermal-infrared SLAM\\Beyond the Visible Spectrum
}
\author{Young-Sik Shin${}^{1}$ and Ayoung Kim${}^{1*}$% <-this % stops a space
\thanks{Y. Shin and A. Kim are with the Department of Civil and Environmental Engineering,
        KAIST, Daejeon, S. Korea \texttt{[youngsik.shin, ayoungk]@kaist.ac.kr}}%
\thanks{This work is supported by a grant from the Korea MOTIE (No.10051867)}%
}

\begin{document}

%\onecolumn
\maketitle
\thispagestyle{empty}
\pagestyle{empty}

\input{abstract}
\input{introduction}
\input{literature}
\input{calibration}
\input{thermalslam}
\input{results}
\input{conclusion}

%\section*{ACKNOWLEDGMENT}

\balance
\bibliographystyle{IEEEtran}
\bibliography{string-short,ral2019-yshin}

\end{document}

%% file: abstract.tex
\begin{abstract}

In this paper, we propose a thermal-infrared \ac{SLAM} system enhanced by sparse
depth measurements from \ac{LiDAR}. Thermal-infrared cameras are relatively
robust against fog, smoke, and dynamic lighting conditions compared to RGB
cameras operating under the visible spectrum. Due to the advantages of
thermal-infrared cameras, exploiting them for motion estimation and mapping is
highly appealing. However, operating a thermal-infrared camera directly in
existing vision-based methods is difficult because of the modality difference.
This paper proposes a method to use sparse depth measurement for 6-DOF motion
estimation by directly tracking under 14-bit raw measurement of the thermal
camera. In addition, we perform a refinement to improve the local accuracy and
include a loop closure to maintain global consistency. The experimental results
demonstrate that the system is not only robust under various lighting conditions
such as day and night, but also overcomes the scale problem of monocular
cameras. The video is available at \href{https://youtu.be/oO7lT3uAzLc}{https://youtu.be/oO7lT3uAzLc}.

\end{abstract}

%% file: introduction.tex
\section{Introduction}
\label{sec:intro}

Ego-motion estimation and mapping is a crucial factor for an autonomous vehicle.
Much of the research in robotics has focused on imaging sensors and LiDAR
\cite{cforster-2014, rmurartal-2015, jzhang-2017a} for navigating
through environments without \ac{GPS}. Conventional RGB cameras operating under
the human visible spectrum hinder operating in a challenging environment such as
fog, dust, and complete darkness. Recently, thermal-infrared cameras have been
highlighted by studies for their perceptual capability beyond the visible
spectrum and robustness to environmental changes.

Despite the outperformance against conventional RGB cameras, leveraging a
thermal-infrared camera into previous visual perception approaches is a
challenge in several respects. In particular, the distribution of data is
usually low contrast because the temperature range sensed by a thermal-infrared
camera is wider than the ordinary temperature range of daily life. In addition,
due to the nature of thermal radiation, observable texture information in the
human visible spectrum is lost and contrast is low in the images. To utilize
previous approaches, a strategy of rescaling thermal images is used in
\cite{tmouats-2015, jpoujol-2016, abeauvisage-2016, cpapachristos-2017}. There
are roughly two ways to rescale 14-bit raw radiometric data. First is to apply
the histogram equalization technique using the full range of radiometric data
present in the scene. However, performing a rescale using the entire range in
this way causes a sudden contrast change if a hot or cold object suddenly
appears. The change in contrast causes problems for both direct methods that
require photometric consistency and feature-based methods that require
thresholding for feature extraction. The second method is to manually set the
range in which the rescale technique operates. Although there is a limitation to
changing the range depending on the operating environment, this method maintains
the photometric consistency according to the thermal condition. When a long-term
operation is required, image degradation appears due to accumulated sensor noise
and eventually deteriorates the performance of the \ac{SLAM}\cite{lchen-2017}.

\begin{figure}[!t]%
  \centering%
  \includegraphics[width=0.96\columnwidth]{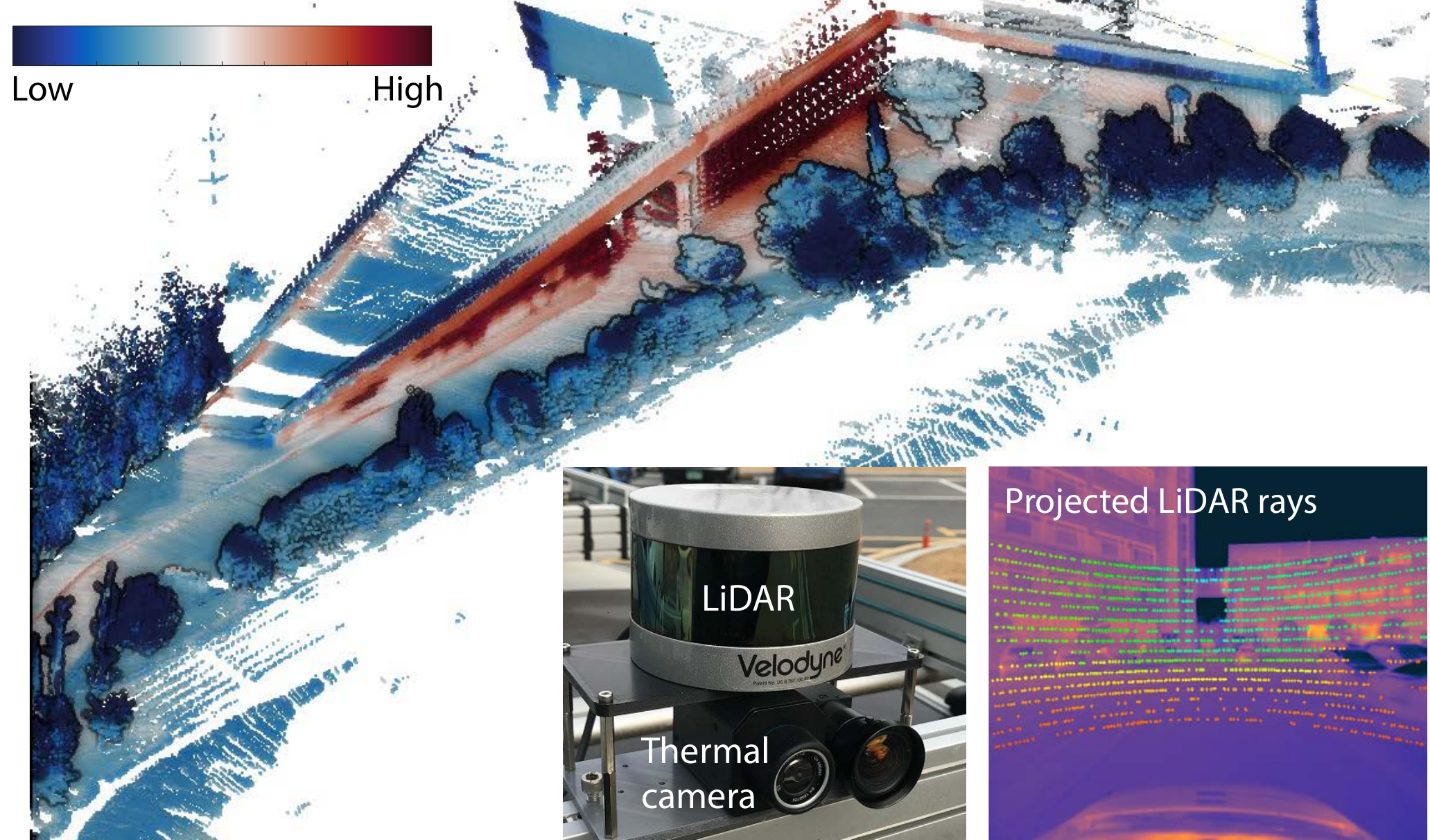}%

  \caption{Sample 3D thermographic mapping result. Left thumbnail image shows
  the configuration of the thermal-lidar sensor system. Right thumbnail is the
  projected LiDAR points on the thermal image. }

  \label{fig:intro}%
\end{figure}

In this paper, we propose a \textit{direct} thermal-infrared \ac{SLAM} system
enhanced by sparse range measurements. The proposed method tightly combines a
thermal-infrared camera with a \ac{LiDAR} to overcome the abovementioned
challenges. Note that as shown in \figref{fig:intro}, only sparse measurement is
available from the \ac{LiDAR}. In addition, this sensor configuration solves the
problem of initialization and scale of monocular SLAM approaches. However,
precise calibration is required to utilize the two tightly coupled sensors.  We
achieve the calibration by utilizing the observable chessboard plane as a
constraint in the thermal-infrared camera.

Inspired by direct methods \cite{ckerl-2013a, cforster-2014, jengel-2017}, the
proposed system tracks the sparse depth provided by \ac{LiDAR} directly on the
14-bit raw thermal image. Unlike previous approaches that scale this 14-bit
image to 8-bit, we utilize the raw temperature in 14-bit without conversion; doing
so eliminates the heuristic preprocessing that rescales the value of the thermal
image to 8-bit. Furthermore, we avoid applying RGB-D sensors based on structured
\ac{IR} patterns and obtain depth measurement that operates regardless of light
condition from day to night.

%% TODO: Our method / Contribution 소개.
The main characteristics of the proposed method can be explained as follows:

\begin{itemize}

%  \item We propose automatic extrinsic calibration method between thermal-infrared
%camera and \ac{LiDAR}. We utlize the constraint the plane-model pair of the pattern
%board detected in camera and \ac{LiDAR}.

  \item We eliminate the heuristic preprocessing of the thermal imaging camera by
  tracking the sparse depth measurement on a 14-bit raw thermal image. Hence the
  high accuracy in the motion estimation stage can be guaranteed without
  geometric error models such as reprojection error of feature correspondences
  or \ac{ICP}.

  \item The proposed method introduces a thermal image based loop-closure to
  maintain global consistency. When calculating the pose constraint, we consider
  the thermographic change from a temporal difference between the candidate
  keyframe and the current keyframe.

  \item Our system completes thermographic 3D reconstruction by assigning
  temperature values to points. The proposed method performs robustly regardless
  of day and night.

  \item Finally, experimental results show that a thermal-infrared camera with a
  depth measurement beyond the visible spectrum ultimately leads to an all-day
  visual SLAM.

\end{itemize}

%% file: literature.tex
\section{Related Works}
\label{sec:relatedworks}

% feature / direct

During the last decade, vision-based motion estimation using consecutive images
has matured into two main streams. One of them is a 6-DOF motion estimation
method that utilizes visual feature correspondence \cite{rmurartal-2015, icvisic-2018}. The other approach, called a direct method,
estimates the ego-motion by minimizing the intensity difference of reprojected
pixels from depth \cite{ckerl-2013a, cforster-2014, jengel-2014, jengel-2017}.
It depends on visually rich information under sufficient lighting conditions.

The robustness of motion estimation under varying illumination is a critical
issue for robotic applications. Due to the physical limitations of the generic
imaging sensor, securing robustness of visual odometry methods under various
lighting conditions (e.g., sunrise, sunset and night) has hardly been achieved
\cite{pkim-2017}. In this respect, utilizing unconventional imaging sensors
operatable under non-visible spectra has drawn attention.

% Thermal Odometry
Among them, thermal-infrared imaging has been highlighted. One approach
introduces a method of visual and thermal-infrared image fusion based on
\ac{DWT}  and applies the representation to the visual odometry framework for
the robustness in \cite{jpoujol-2016}. Another study suggested an odometry
module that uses multi-spectral stereo matching instead of the fusion-based
approach \cite{abeauvisage-2016}. The authors used \ac{MI} to perform
cross-modality matching and temporal matching to improve matching accuracy.
Instead of tightly fusing a multi-spectral sensor, a method of integrating each
sensor's measurement as a map point has also been introduced by
\cite{lchen-2017}. Additionally, using thermal stereo for odometry module has
been proposed for \ac{UAV} in \cite{tmouats-2015}.

% Thermal mapping
In past decades, many studies on the 3D thermographic mapping have been
introduced. Particularly in the remote sensing field, research has been
conducted to integrate aerial thermal imagery and \ac{LiDAR} data for ground
survey and cities modeling \cite{gbitelli-2015, emandanici-2016}. A
thermographic mapping approach utilizing a terrestrial laser scanner has also
been proposed instead of requiring expensive aerial equipment
\cite{dgonzalez-2012}. However, these methods have limitations in mobility
because they require expensive equipment and have significantly lower
portability.

Together with the real-time mobile 3D reconstruction \cite{rnewcombe-2011b,
twhelan-2015}, portable mobile thermographic mapping methods have also been
studied for building inspection \cite{svidas-2013a, svidas-2013b}. Since dense
depth is available from an RGB-D sensor in their application, a geometric error
model such as \ac{ICP} was utilized. Recent studies have used the
thermographic error model to improve robustness. More recently, methods
have been introduced that combine thermographic error models to improve
robustness \cite{szhao-2017, ycao-2018}. However, these methods that heavily
rely on RGB-D sensors and are less suitable for an outdoor environment.

Targetting both indoor and outdoor, we take advantage of the direct visual-LiDAR
SLAM method to our sensor system \cite{yshin-2018} and modify it to the newly
proposed thermal-LiDAR system as shown in \figref{fig:intro}. In this paper, we
summarize modifications and changes toward depth-enhanced direct SLAM for thermal
cameras enabling all-day motion estimation and 3D thermographic mapping. The
detailed topic includes 14-bit loss function for direct method, extrinsic
calibration between a thermal camera and a LiDAR, and temperature bias
correction during a loop-closure.

% thermal in SLAM
% https://www.autonomousrobotslab.com/autonomous-navigation-and-exploration.html
% K. Alexis intiated robotics point of view (localization)

%%SURFEL = mapping

%There are approaches that focus on the 3D reconstruction stage to remove the
%effects of moving objects. \hl{TODO: cite Keller / Elastic Fusion} They are
%registered by fusing only points that are repeatedly observed in consecutive
%frames to surfels. Robustness in dynamic environment is ensured by rendering and
%tracking local 3D surfel maps that are not affected by dynamic elements. However,
%this methodology does not explicitly detect dynamic elements.  In addition, this
%methodology is not scalable if only sparse measurements are available.

%% file: calibration.tex
\section{Automatic Thermal Camera-to-LIDAR Calibration}
\label{sec:calibration}

%Figure
%\begin{figure}[!t]
%\centering
%    \includegraphics[width=0.4\columnwidth]{figs/sensor_unit_annot}
%  \caption{Sensor system. \hl{change required.}}
%  \label{fig:sensor_system}
%\end{figure}
%Figure

In this section, we describe the estimation of the relative transformation
$\mathbf{T}^c_v$ between the thermal camera coordinate frame $\{c\}$ and the
\ac{LiDAR} coordinate frame $\{v\}$ where two sensors are assembled as shown in
\figref{fig:intro}. Given the initial transformation by hardware design, we
optimize to obtain relative transformation between two sensors in terms of the
extrinsic calibration parameters. The calibration accuracy is critical in the
proposed method as a small calibration error may yield a large drift in the
entire trajectory estimation.

Existing methods often aimed at estimating the relative pose between the RGB
camera and the range sensor \cite{ageiger-2012, gwyeth-2010, qzhang-2004}. Even
though the geometry of the thermal camera is similar to that of an RGB camera,
these methods do not directly apply to our sensor system because the visible
spectrum is completely different. The existing approaches for thermal-LiDAR
extrinsic calibration \cite{jlussier-2014, ychoi-2018} require dense range data
together with a user intervention for data selection. Instead, we propose an
automated extrinsic calibration method that utilizes an observable chessboard
pattern in the spectrum of a thermal-infrared camera, while requiring little
user intervention. Using a pattern board that is visible to a thermal camera,
we leverage a pattern board plane obtained from a data stream
instead of using multiple planes in a single scene.

%----------------------------------------------------------------------------%
\subsection{Pattern Board}

Pattern boards printed on generic paper are widely used in RGB camera
calibration. However, they are only effective in the visible spectrum and
reveal near-uniform radiation for a thermal camera. Hence, several chessboard
patterns have been introduced to obtain intrinsic parameters of thermal cameras
\cite{ychoi-2018}. Inspired by these methods, we used a pattern board that utilizes
a \ac{PCB}. Because a conductive copper surface appears as a white region
in a thermal camera image, we can use the conventional method of chessboard
detection. This pattern board was used for both intrinsic and extrinsic
calibration of thermal cameras. % \hl{PHOTO PATTERN BOARD}

%----------------------------------------------------------------------------%
\subsection{Plane segmentation}

The purpose of segmentation in the calibration phase is to automatically find a
pattern board and plane parameter from two sensors. The proposed method assumes
that the initial extrinsic parameters can be obtained by hardware design.

Once the intrinsic parameters of the thermal camera and the size of the pattern
have been given, we can easily estimate the chessboard pose in the camera
coordinate system using existing methods \cite{rhartley-2003}. Then, we detect
the plane of the pattern board observed in the \ac{LiDAR} coordinate system.
Using the initial extrinsic parameter, we project the 3D points of the
\ac{LiDAR} coordinate system onto the thermal image. We use the plane-model
based RANSAC algorithm using only those existing on the pattern board. As a
result, we estimate plane models from the sparse points in the \ac{LiDAR}
coordinate system.

%----------------------------------------------------------------------------%
\subsection{Initial Parameter Estimation}

% Q. auto? or manually select three?
The process of estimating extrinsic parameters between two sensors was inspired
by \cite{ageiger-2012}. Unlike \cite{ageiger-2012}, who utilized single shot
data, we exploit a temporal data stream.  Given a plane data pair from a LiDAR
and a thermal camera, the relative transformation between the sensors is
obtained using the geometric constraints of the plane model. Since the proposed
method uses a temporal data stream in which a single chessboard is observed, the
information from plane models, $ax+by+cz+d=0$, oriented in a similar direction
becomes redundant.  At least three plane models are required to calculate the
transformation, and all possible combinations of selected indexes are
precomputed. Among them, the algorithm selects three plane pairs ($p_a,p_b,p_c$)
from the thermal camera to maximize the following probability:
\begin{equation}
  \small
  p(s_{p_a},s_{p_b},s_{p_c}) = \frac{1}{Z}\exp(-\mathbf{n}_{p_a}^T\mathbf{n}_{p_b}-\mathbf{n}_{p_b}^T\mathbf{n}_{p_c}-\mathbf{n}_{p_a}^T\mathbf{n}_{p_a}),
\end{equation}
where $Z$ is the normalization factor, $s_{p_a},s_{p_b},s_{p_c}$ are the indices of the
three selected plane pairs, and $\mathbf{n}$ is the normal of the plane.
Intuitively, the greater the difference between the normals of the three planes,
the higher the probability. Given three plane pairs, we solve the relative
rotation of the two sensors via the equation below:

\begin{equation}
  \argmax_{\mathbf{R}^v_c}\sum_{i\in\{p_a,p_b,p_c\}}{\mathbf{n}^i_v}^T\mathbf{R}^v_c\mathbf{n}^i_c=\mathbf{V}\mathbf{U}^T,
\end{equation}
where the subscripts $c$ and $v$ indicate the coordinates of the camera and \ac{LiDAR}.
By performing \ac{SVD} of the covariance matrix
$\sum_{i}{\mathbf{n}^i_c{\mathbf{n}^i_v}^T}=\mathbf{U}\Sigma\mathbf{V}^T$, the
rotation matrix $\mathbf{R}^v_c$ estimation that maximizes the alignments of
the normal vectors is calculated:
\begin{equation}
\argmin_{\mathbf{t}}\sum_{i\in\{p_a,p_b,p_c\}}{\mathbf{n}^i_v}^T(\mathbf{R}^v_c\mathbf{p}^i_c+\mathbf{t}^v_c)+d_v.
\end{equation}
Next, we obtain a translational vector that minimizes the distance between a
point found in the camera and a plane-model extracted from \ac{LiDAR}.

%Figure
\begin{figure}[!t]
\centering
    \includegraphics[width=0.9\columnwidth]{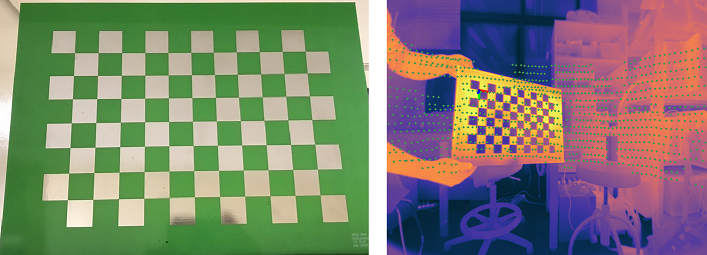}%
  \caption{Pattern board and the calibration result. The green
points represent lidar measurements projected onto a thermal image.}
  \label{fig:calib_result}
  \vspace{-4mm}
\end{figure}
%Figure

%----------------------------------------------------------------------------%
\subsection{Refinement}

Following the previous section, the initial parameters were calculated
using the normal and plane-to-point distances of the planes observed at each
sensor. This method can be computed quickly in a closed form, but is less
accurate and requires additional refinement. In this work, we use a gradient
descent method to refine the relative transformation that minimizes the
cost as following:
\begin{equation}
  \argmin_{\mathbf{R^v_c}, \mathbf{t}}\sum_{i}{\mathbf{n}^i_v}^T(\mathbf{R}^v_c\mathbf{p}^i_c+\mathbf{t}^v_c)+d_v.
\end{equation}
At this stage, we use all plane data pairs obtained from the segmentation
procedure mentioned in the previous section. \figref{fig:calib_result} shows the
calibration results by the proposed method. The points in the \ac{LiDAR} coordinate
system are projected into thermal camera images.

%% file: thermalslam.tex
\section{Sparse Depth Enhanced Direct Thermal-Infrared SLAM}
\label{sec:thermal_SLAM}

\begin{figure}[!t]
\centering
  \centering%
  \includegraphics[width=0.9\columnwidth]{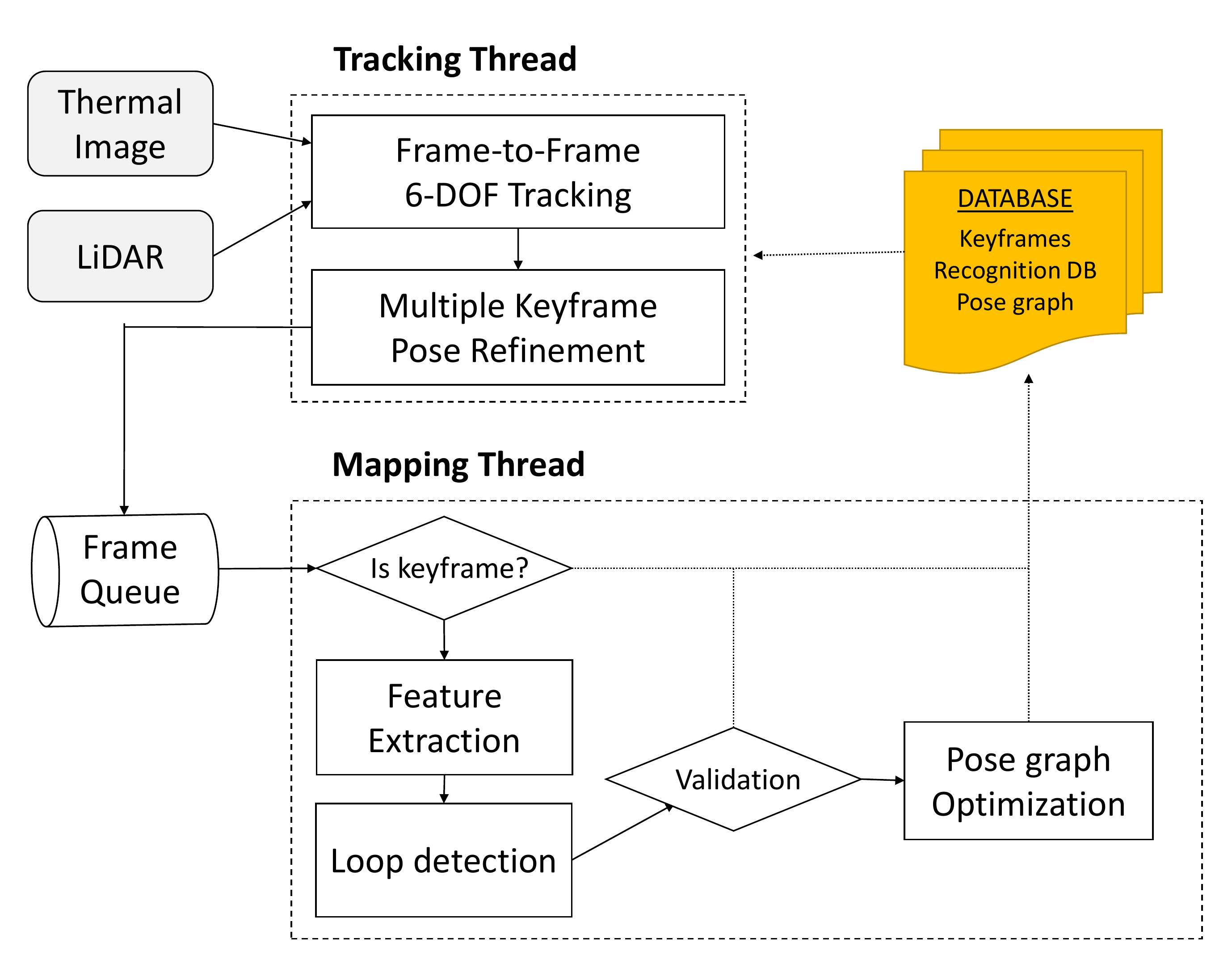}%
  \caption{Block diagram of the direct thermal-infrared SLAM with sparse depth.
    The proposed method consists of a tracking module for estimating the pose of
    each frame and a loop closing module for global consistency.}
  \label{fig:overview}
  \vspace{-4mm}
\end{figure}

Given an extrinsic calibration, we can project LiDAR points onto a thermal
image frame. \figref{fig:overview} provides the pipeline of the thermal-infrared
SLAM. The proposed thermal-infrared \ac{SLAM} method focuses on (\textit{i})
accurate 6-DOF tracking of the direct method and (\textit{ii}) a reliable loop
detection method based on the visual features. The thermal-infrared SLAM consists
of a tracking thread for estimating the 6-DOF motion of the input frame and a
mapping thread for maintaining global consistency. We assume that the \ac{LiDAR} and
the thermal camera are already synchronized. Because depth measurements from
\ac{LiDAR} are available even in a sparse form, we are able to exploit the
measurement instead of triangulating the corresponding points from the
thermal-infrared camera.

The tracking thread \textit{directly} estimates the relative motion by tracking
the points of the previous frame and then performs pose refinement from multiple
keyframes to improve local accuracy. We utilized the point sampling method for
computational efficiency similar to \cite{yshin-2018}. By tracking the
\ac{LiDAR} depth measurement associated with the 14-bit raw thermal image, we
can easily take advantage of the direct method. Note that we exploit only the
sparse points within the camera \ac{FOV}, suggesting potential application to a
limited FOV LiDAR (e.g., solid-state LiDAR).

Lastly, the mapping thread aims to maintain global consistency. We utilized a
bag-of-word-based loop detection when the current frame is revisited.  However,
the thermal images lack detailed texture information unlike the RGB images in
the visible spectrum, thus naive implementation results in repetitive frequently
detected pattern features and inaccurate loop candidates. Hence, we enforce the
geometric verification to avoid adding incorrect loop constraints to the
pose-graph.

%--------------------------------------------------------------------------%
\subsection{Notation}

%Figure
%Figure

A temperature image $\mathcal{I}_i$ and a sparse point cloud
$\mathbf{\mathcal{P}}^i_v$ are provided via synchronization. Using the extrinsic
parameters obtained in the previous section, we obtain the transformed point
cloud $\mathbf{\mathcal{P}}^i_c$ in the camera coordinate. Then, each point
$\mathbf{p}^k_c=(x,y,z)^\top\in\mathbb{R}^3$ is represented as a pixel
$\mathbf{u_k}=(u,v) \in \mathbb{R}^2$ on the image coordinate system through the
projection function $\pi(\mathbf{p}_k)$ via
\begin{equation}
  \mathbf{u_k} = \pi(\mathbf{p}_k) = \left(\frac{x f_x}{z} + c_x, \frac{y f_y}{z} + c_y \right).
\end{equation}

The transformation matrix $\mathbf{T}_n \in SE(3)$ on the world coordinate
represents the pose of each frame and the relative pose between $\mathcal{F}_n$
and $\mathcal{F}_{m}$ is denoted by
$\mathbf{T}_m^n=\mathbf{T}_{n}^{-1}\mathbf{T}_{m}$. This relative pose is also
used for the coordinate transformation of points as follows:
\begin{equation}
  \label{eq:proj}
  \mathbf{p}_n = \mathbf{T}_m^n\mathbf{p}_m =
  \begin{bmatrix}
    \mathbf{R}_m^n & \mathbf{t}_m^n \\
    \mathbf{0} & 1
  \end{bmatrix}
  \mathbf{p}_m,
\end{equation}
where $\mathbf{p}_m$ represent points on the frame $\mathcal{F}_m$, $\mathbf{R}_m^n \in
SO(3)$ and $\mathbf{t}_m^n \in \mathbb{R}^3$ are a rotation matrix and
translation vector. Lie algebra elements $\mathbf{\xi} \in \mathbb{R}^6$ are
used as linearized pose increments. The twist $\xi= [\mathbf{w}, \mathbf{v}]$
consist of an angular and linear velocity vector. The updated transformation
matrix using the increment can be calculated by exponential mapping from Lie
algebra to Lie group: $se(3) \times SE(3) \rightarrow SE(3)$, \ie,

\begin{equation}
  \label{eq:Tnm}
  \mathbf{\bar{T}}_m^n = \exp(\hat{\xi}) \mathbf{T}_m^n.
\end{equation}

%--------------------------------------------------------------------------%
\subsection{Tracking}

Inspired by recent direct methods \cite{cforster-2014,jengel-2017}, our tracking
method utilizes a thermographic error model. Instead of a photometric error
model that utilizes the intensity of the image, we use the 14-bit temperature
value provided by the thermal-infrared camera. Additionally, we apply a patch
$\Omega$ of the sparse pattern to ensure robustness as in \cite{jengel-2017}.
Although this patch-based approach requires more computation, it provides robust
results for image degradation such as motion blur or noise.

In the tracking process, the points $\mathbf{p}_m$ of the $m^\text{th}$ frame
can be projected onto the image $\mathcal{I}_n$ of the current frame
$\mathcal{F}_n$ by the initial relative pose. Thermographic residuals are
expressed as the difference in temperature value as shown below
\begin{equation}
  \label{eq:rp}
  r(\mathbf{p}_m) = \mathcal{I}_n(\pi(\mathbf{T}_m^n\mathbf{p}_m))-\mathcal{I}_m(\pi(\mathbf{p}_m)).
\end{equation}
Finally, our objective function for tracking is defined as follow:
\begin{equation}
  \label{eq:etrack}
  \mathit{E}_{track}=\sum_{\mathbf{p}\in \Omega} w(r(\mathbf{p})) (r(\mathbf{p}))^2,
\end{equation}
where $\Omega$ is the patch of the sparse pattern and the function $w(\cdot)$
is a weight function based on the t-distribution, which is calculated iteratively
and recursively as shown in \cite{ckerl-2013a, yshin-2018}:
\begin{equation}
  \label{eq:w}
  w(r) = \frac{\nu+1}{\nu + (\frac{r}{\sigma_r})^2},
\end{equation}
where $\nu$ is the degree of freedom, which was set to five, and the variance of
residual $\sigma_r$ is iteratively estimated while performing the Gauss-Newton
optimization algorithm. The effect of weighting is reported in
\cite{ckerl-2013a}. Finally, a coarse-to-fine scheme is used to prevent
large displacements that cause a local minimum of the cost function.

%--------------------------------------------------------------------------%
\subsection{Local Refinement}

In the previous section, the tracking process estimates 6-DOF relative motion by
frame-to-frame. When the time interval between frames is short, tracking
loss and potential local drift inevitably occur. To reduce this drift and
improve local accuracy, we perform a multiple keyframe-based refinement. Once the
frame-to-frame tracking process is completed, we perform optimization using the
recent keyframes. The main difference with the tracking process is that
optimization is performed on map coordinates based on multiple keyframes.

Given that the poses of the keyframes are defined in the map coordinate, we update
the current frame pose using the relative motion obtained from the tracking
process prior to the refinement phase using the cost below:

\begin{equation}
  \label{eq:ewin}
  \mathit{E}_{refine}= \sum_{\mathcal{KF}_i \in \mathbf{W}_{kf}} \sum_{\mathbf{p}_k\in \mathbf{P}_{\mathbf{i}}} \sum_{\mathbf{p}_k \in \Omega} w(r(\mathbf{p}_k)) (r(\mathbf{p}_k))^2,
\end{equation}
 where $\mathbf{W}$ represents the window of multiple keyframes, $\mathbf{p}$
indicates the points of the keyframe and $\Omega$ means the patch of the sparse
pattern. The weight function $w(\cdot)$ uses t-distribution, as described in
the previous section. The residual function $r(\cdot)$ is defined using the raw
values of the keyframe and the current frame as:
\begin{equation}
  \label{eq:r_refine}
  r(\mathbf{p}_m) = \mathcal{I}_f(\pi(\mathbf{T}_{f}^{-1}\mathbf{T}_{kf}\mathbf{p}_{k}))-\mathcal{I}_{kf}(\pi(\mathbf{p}_{k})).
\end{equation}
Lastly, the equation \eqref{eq:ewin} is optimized with respect to the pose of
the current frame using the Gauss-Newton method.

%--------------------------------------------------------------------------%
\subsection{Loop Closing}

\subsubsection{Loop Detection}

The inferred thermal odometry estimation may deteriorate the global consistency
due to the accumulation of small drifts and errors. We handle this issue by
relying on a loop closure that takes advantage of the constraints between the
revisited place and the current frame. In the mapping thread, detection of the
revisiting area depends on \texttt{DBoW}, a place recognition method based on
the bag-of-binary-words \cite{glopez-2012}.

Each time a new keyframe is created, we extract the ORB feature from the thermal
image. Instead of directly utilizing the 14-bit raw data for loop closure, we
chose a strategy that performs equalization through a linear operation in a
fixed range of ordinary temperature ($0^\circ$ to $30^\circ$). Because only
the simple linear operation is required, we can expect the applicability of the feature
extraction methodology to 14-bit raw data.

Next, we calculate the bag-of-words vectors then calculate the normalized
similarity score from the keyframe database.

\begin{equation}
   \label{eq:sim_score}
   \eta(\mathbf{v}_c,\mathbf{v}_q) = \frac{s(\mathbf{v}_c,\mathbf{v}_q)}{s(\mathbf{v}_c,\mathbf{v}_1)}
\end{equation}
 where $\mathbf{v}_q$ is the BoW vector of the previous keyframes. To
calculate the similarity score $s(\mathbf{v}_c,\mathbf{v}_q)$, the $L_1$ distance
is used as shown below:
\begin{equation}
   \label{eq:sim_vec}
   s(\mathbf{v}_c,\mathbf{v}_q) = 1-\frac{1}{2}
   \left|
   \frac{\mathbf{v}_c}{|\mathbf{v}_c|} - \frac{\mathbf{v}_q}{|\mathbf{v}_q|}
   \right|
\end{equation}
Since keyframes are added sequentially while excluding keyframes with a
timestamp close to the current frame from the loop candidate, an additional
check on the ratio of common words is performed to prevent false positives.

\begin{figure}[!t]%
  \centering%
  \def\width{0.4\columnwidth}%
  \subfigure{%
    \includegraphics[width=\width]{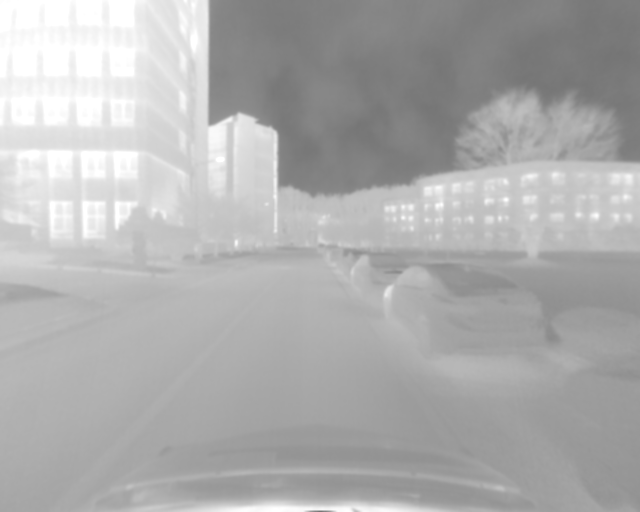}%
    \label{F:query1}%
  }\hfil%
  \subfigure{%
    \includegraphics[width=\width]{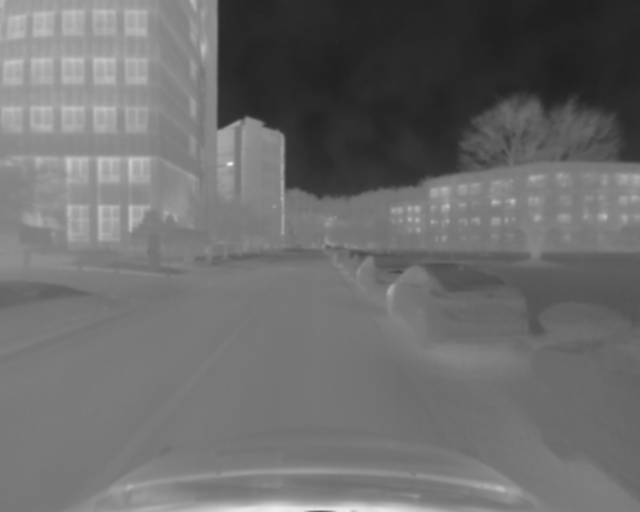}%
    \label{F:db1}%
  }
  %\\
  \subfigure{%
    \includegraphics[width=\width]{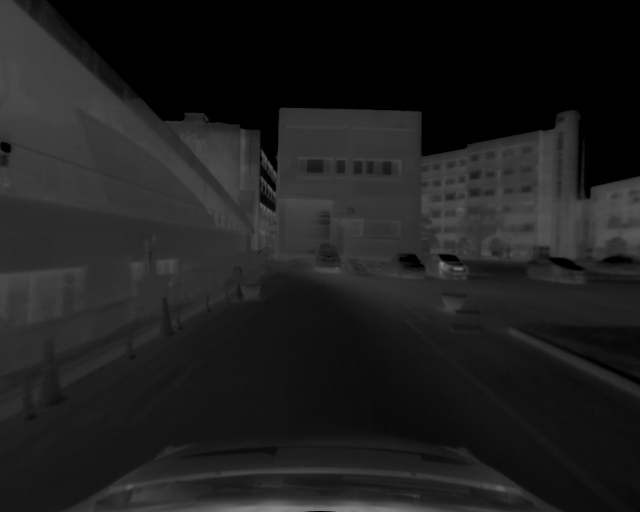}%
    \label{F:query2}%
  }\hfil%
  \subfigure{%
    \includegraphics[width=\width]{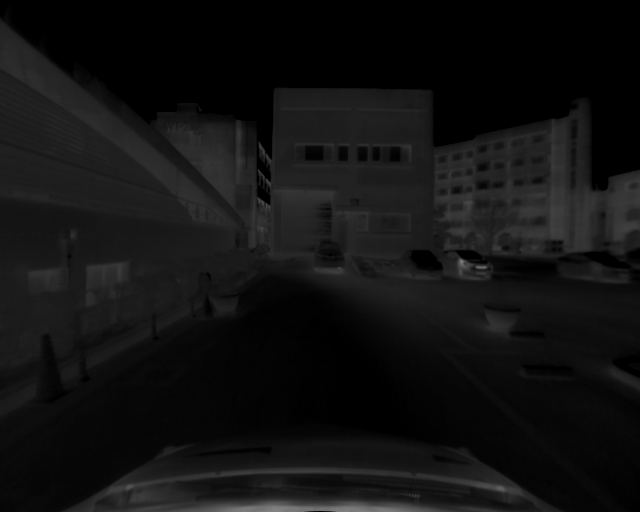}%
    \label{F:db2}%
  }

  \caption{Example of inconsistency due to temporal difference between loop candidate keyframe and query frame. The left column is the query frame and the right image is the image of the revisited keyframe found. Note the image intensity (temperature) difference over time.}

  \label{fig:temporal_diff}%
  \vspace{-4mm}
\end{figure}

\subsubsection{Loop Correction}

When a candidate keyframe $\mathcal{F}_{kf}$ is detected for the loop-closure,
we perform cross-validation by calculating each relative pose using the points
owned by the current frame $\mathcal{F}_{c}$ and the candidate keyframe
$\mathcal{F}_{kf}$.  Given a current frame $\mathcal{F}_{c}$ and a candidate
keyframe $\mathcal{F}_{kf}$, we estimate two relative motions
$\mathbf{T}^{kf}_{c}$ and $\mathbf{T}^{c}_{kf}$.

Another notable phenomenon in operating thermal visual odometry outdoors is the
gradual shift in the temperature during a mission. When a significant temporal
difference between the two frames occurred, the consistency of the two images
collapsed. As shown in \figref{fig:temporal_diff}, unlike RGB images that rarely
change within an hour, we notice that the temperature may reveal substantial
bias over a relatively short period of time. Therefore, the conventional
relative motion estimation using the residual model as in RGB images is
aggravated. To cope with this difference in the thermal images, we used an
affine illumination model from \cite{hjin-2001}.
\begin{equation}
  \label{eq:rp}
  r(\mathbf{p}_{ref}) = (a\mathcal{I}_n(\pi(\mathbf{T}_{ref}^n\mathbf{p}_{ref}))+b)-\mathcal{I}_{ref}(\pi(\mathbf{p}_{ref})).
\end{equation}
Above, the residual model applies to \eqref{eq:etrack} and can be solved iteratively.
However, the parameters $a$ and $b$ for the affine illumination model are known
to react differently to outliers with the 6-DOF motion parameter $\xi$
from \cite{jengel-2015}. We applied an alternative optimization method to
disjointly solve $a$ and $b$ in an iterative optimization process.
\begin{equation}
  \label{eq:cross_val}
  \norm{\log{\left(\mathbf{T}_{kf}^{c}\mathbf{T}_{c}^{kf}\right)}} < \epsilon
\end{equation}
When two relative poses are obtained, we lastly check the consistency between
the two estimates prior to adding the loop-closure to the pose-graph. If the
difference between the two relative pose estimates is smaller than the threshold
$\epsilon$ in \eqref{eq:cross_val}, a loop constraint is added to the pose graph
and the correction is performed.

%% file: results.tex
\section{Experimental Results}
\label{sec:experiment}

%FIGURE
%\begin{figure}[!t]
%  \centering
%  \begin{minipage}{0.71\columnwidth}
%  \subfigure[Vehicle equipped with sensor system]{%
%    \includegraphics[width=0.99\textwidth] {figs/vehicle_a}
%    \label{fig:vehicle}
%  }
%  \end{minipage}
%  \begin{minipage}{0.275\columnwidth}
%  \subfigure[Thermal-\ac{LiDAR}]{%
%   	\includegraphics[width=0.99\textwidth] {figs/vehicle_b}
%   	\label{fig:tlidar}
%  }\\
%  \subfigure[VRS-GPS]{%
%   	\includegraphics[width=0.99\textwidth] {figs/vehicle_c}
%   	\label{fig:vrs_gps}
%  }\\
%  \end{minipage}
% \caption{Sensor configuration for the experiments. \subref{fig:vehicle} shows
%the vehicle with the thermal-camera and \ac{LiDAR} used in the experiment.
%\subref{fig:tlidar} indicate the thermal-\ac{LiDAR} sensor. The visible camera
%shown in (b) is for data acquisition only and (c) is VRS-GPS for the ground truth.}
%\label{fig:equipmment}
%\end{figure}
%FIGURE

The proposed method was validated using a sensor system with rigidly coupled
LiDAR and a thermal-infrared camera as shown in \figref{fig:intro}. The detailed
specifications of the sensors that make up the system are shown in
\tabref{tl:sensor_spec}. An A65 thermal camera from FLIR was used in the
experiment. The camera senses the spectral range in the \ac{LWIR} region and
provides 14-bit raw images with a $90^\circ$ of horizontal \acs{FOV}.

%TABLE
\begin{table}[!h]
  \centering
  \caption{Specifications of thermal camera and \ac{LiDAR}}
  \scriptsize
  \label{tl:sensor_spec}

    \begin{tabular}{c|lll}
    \hline
    Type                                                     & \multicolumn{1}{c}{Manufacturer} & \multicolumn{1}{c}{Model} & \multicolumn{1}{c}{Description}                                                                                                             \\ \hline
    \begin{tabular}[c]{@{}c@{}}Thermal\\ Camera\end{tabular} & FLIR                             & A65                       & \begin{tabular}[c]{@{}l@{}}$90^\circ\times69^\circ$ FOV with 7.5mm lens\\ 7.5 $\sim$13 um spectral range (LWIR)\\ 14-bit 640x512 resolution @ 30Hz \\ $-40^\circ$ to $550^\circ$ temperature range \end{tabular} \\ \hline
    3D \ac{LiDAR}                                            & Velodyne                         & VLP-16                    & \begin{tabular}[c]{@{}l@{}}16 channel $360^\circ$ FOV @ 5$\sim$20Hz\\ measurement range up to 100m\end{tabular}                                      \\ \hline
    \end{tabular}
\end{table}
%TABLE

%FIGURE
\input{fig_img_samples.tex}
%FIGURE

\begin{figure*}[!t]
    \centering
    \def\width{0.3\textwidth}%
    \subfigure[Seq \texttt{9}]{%
        \includegraphics[width=\width]{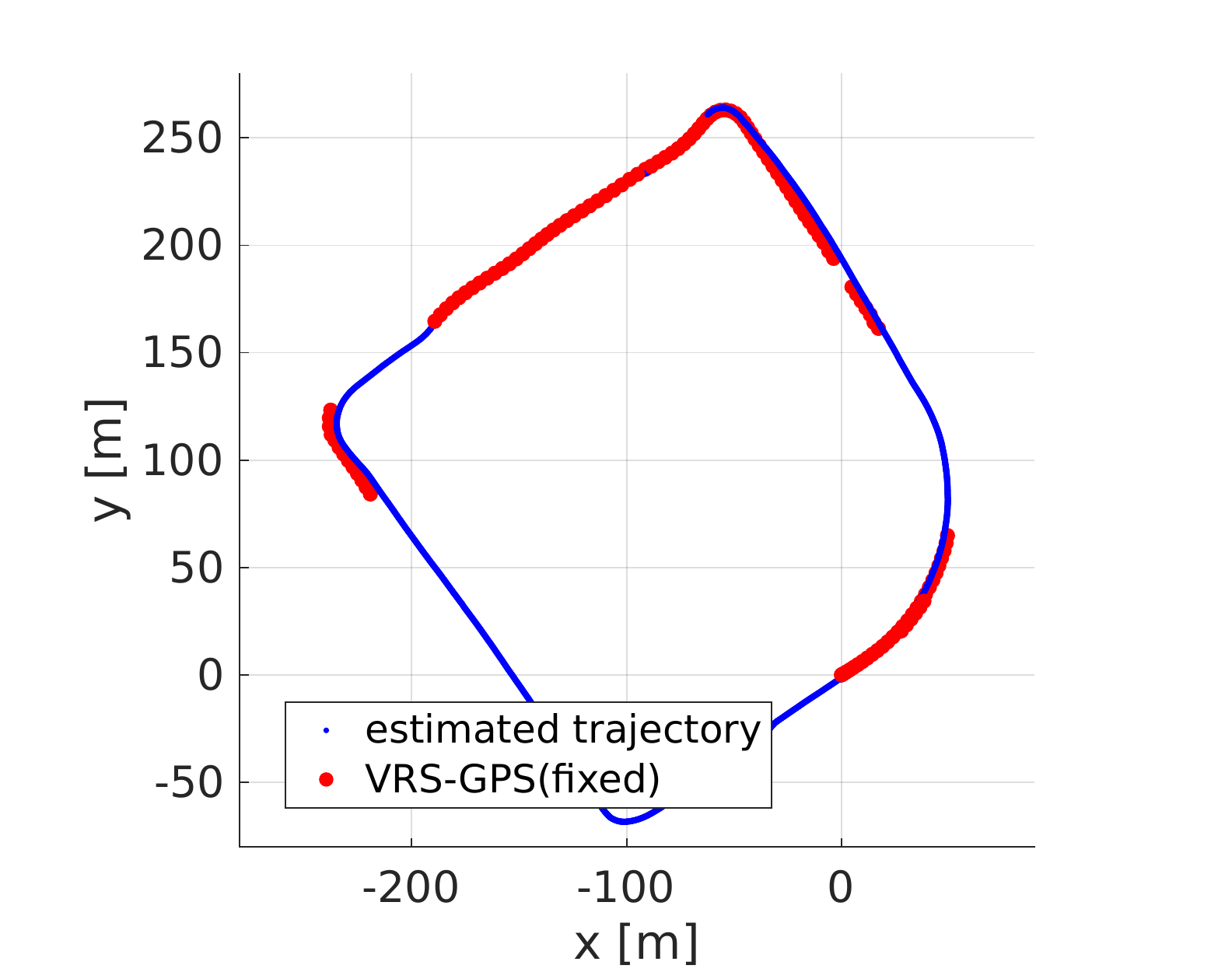}
        \label{fig:seq9_traj}
    }\hfil
    \subfigure[Seq \texttt{11}]{%
        \includegraphics[width=\width]{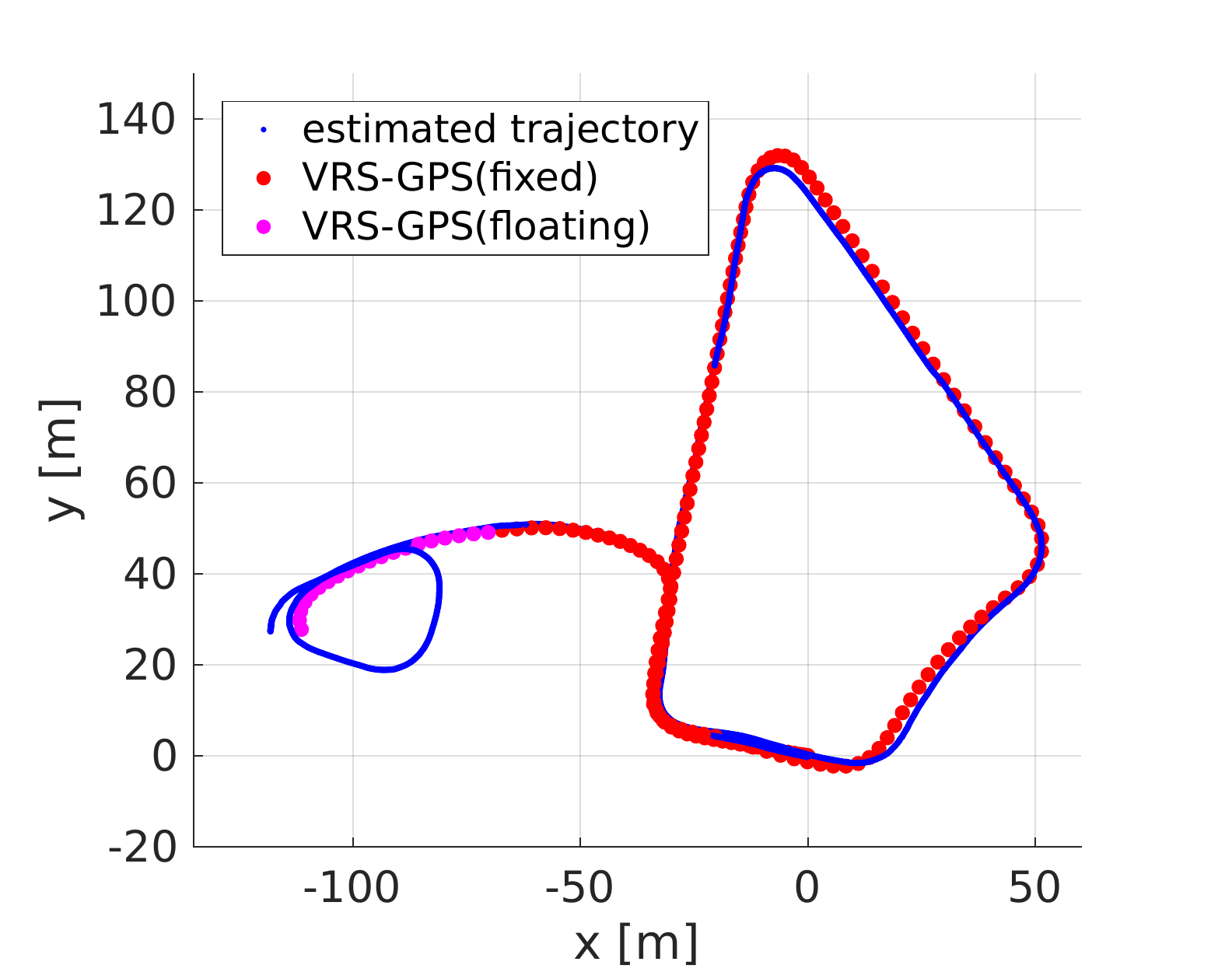}
        \label{fig:seq11_traj}
    }\hfil
    \subfigure[Seq \texttt{14}]{%
        \includegraphics[width=\width]{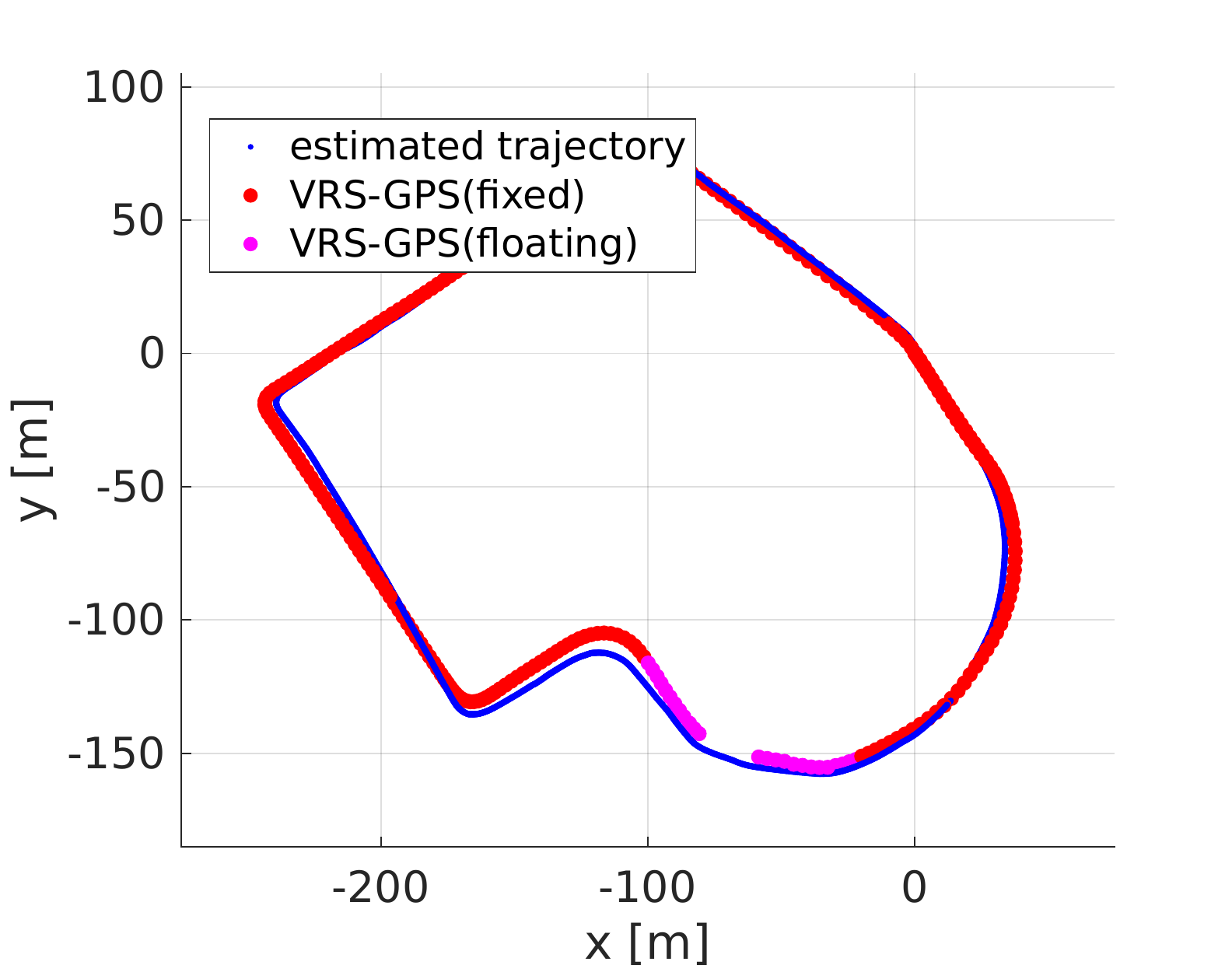}
        \label{fig:seq14_traj}
    }\\
    \subfigure[Map \texttt{9}]{%
        \includegraphics[width=\width]{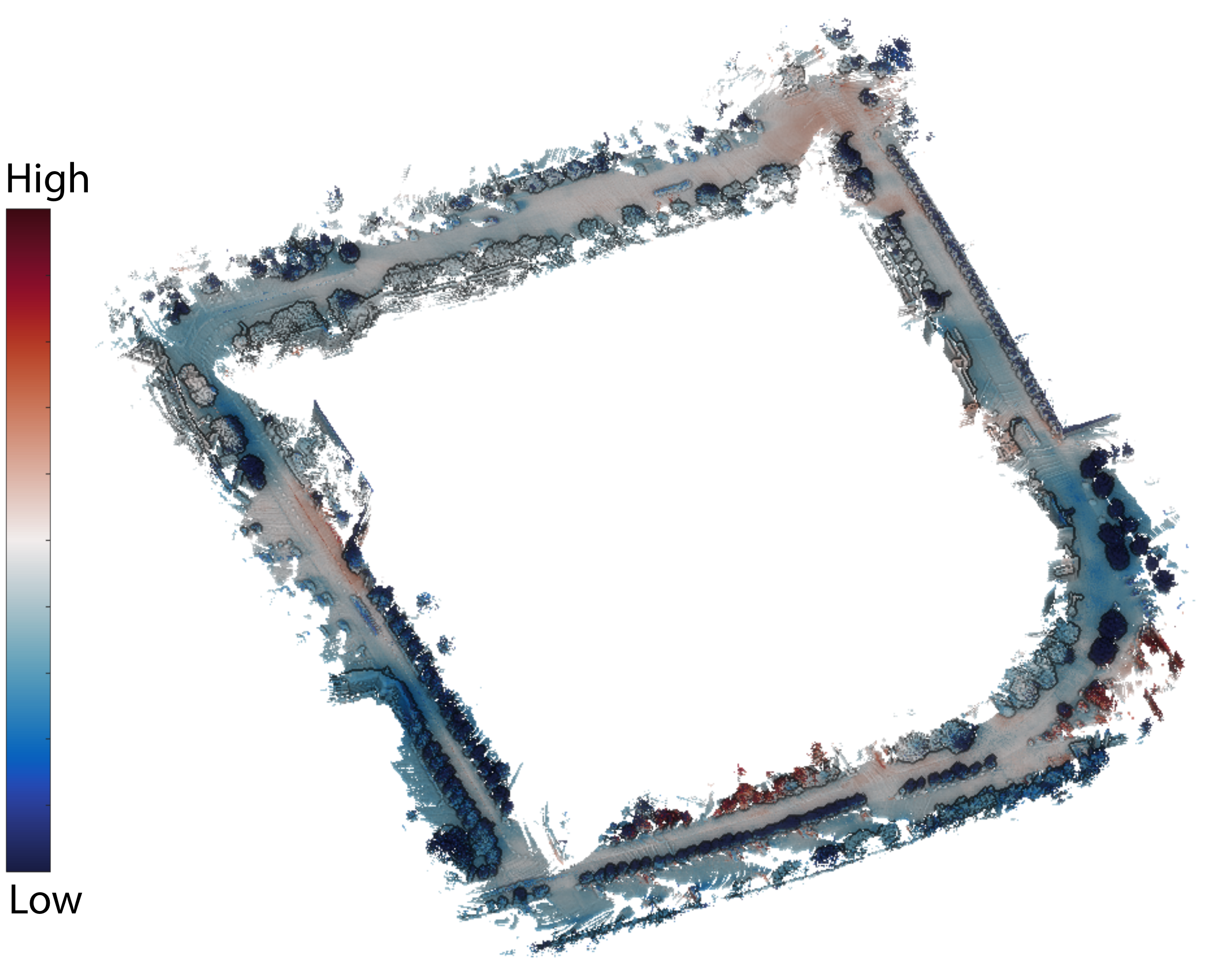}
        \label{fig:thermal_map1}
    }\hfil
    \subfigure[Map \texttt{11}]{%
        \includegraphics[width=\width]{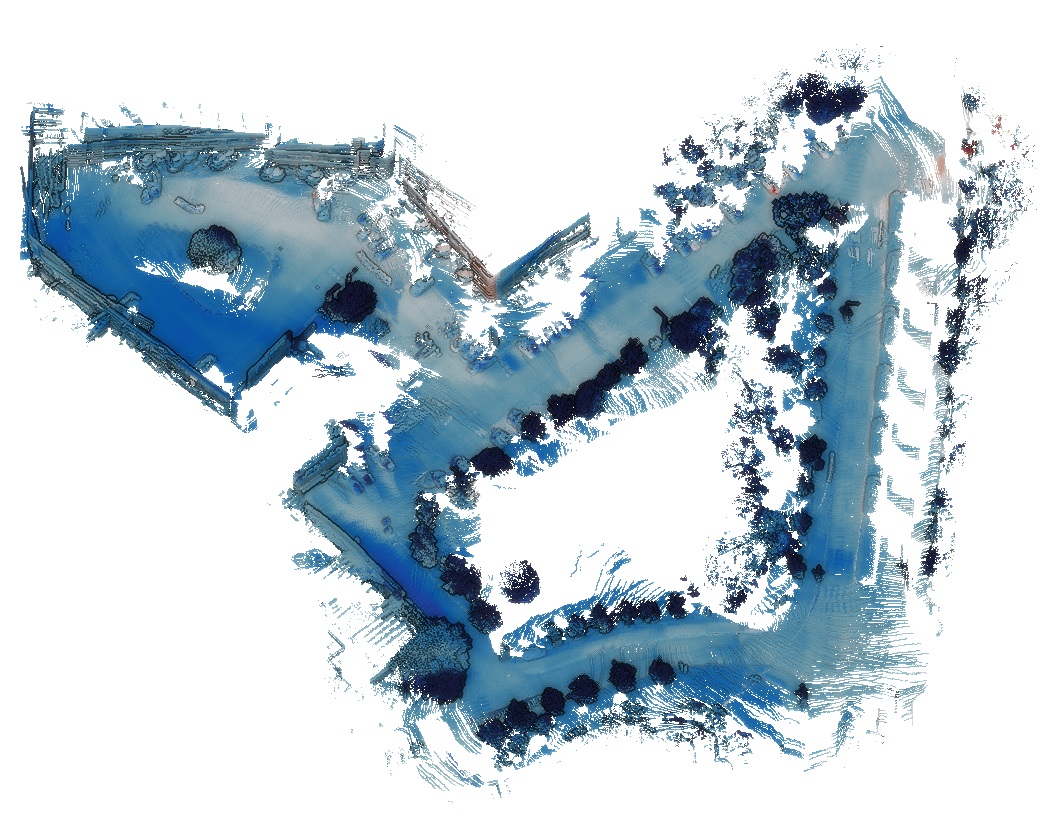}
        \label{fig:thermal_map2}
    }\hfil
    \subfigure[Map \texttt{14}]{%
        \includegraphics[width=\width]{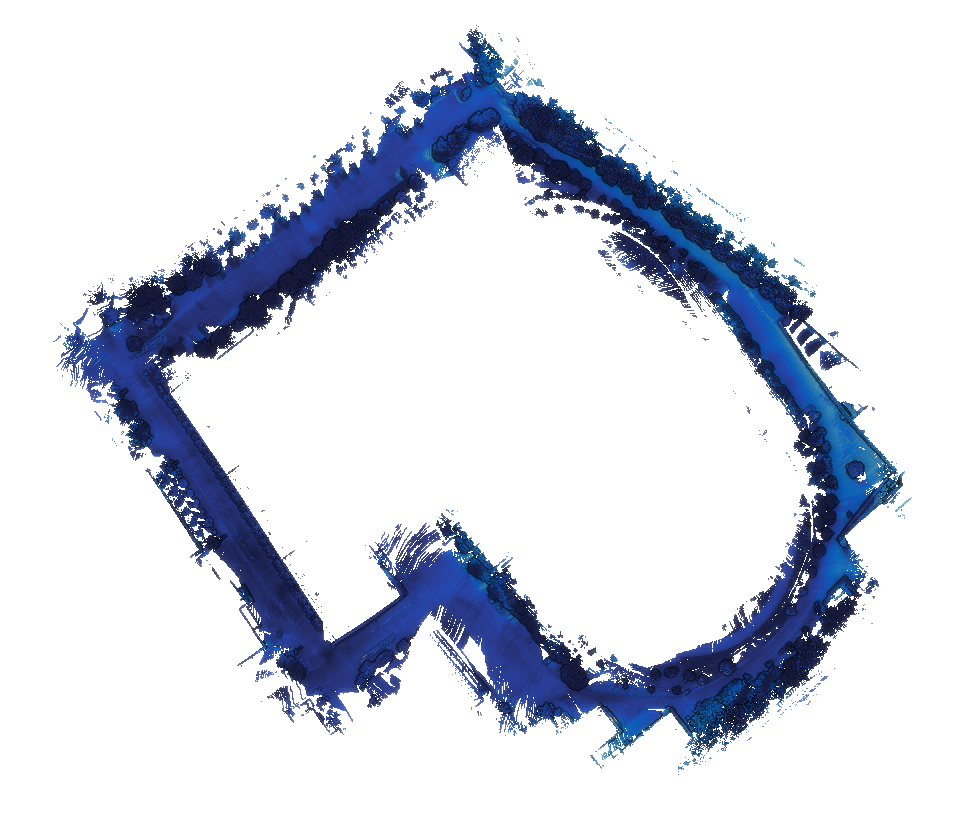}
        \label{fig:thermal_map3}
    }

    \caption{Result of estimated trajectory and 3D temperature mapping.
    \subref{fig:seq9_traj}-\subref{fig:seq14_traj} the estimated trajectory and
    VRS-GPS position information. Regions lacking VRS-GPS accuracy are excluded.
    \subref{fig:thermal_map1}-\subref{fig:thermal_map3} the temperature range is
    colored from 0 to 30 degrees. Note that the night-time temperature is close
    to $0^\circ$, so it represents a dark blue color. We used \texttt{balance}
    colormap from \texttt{cmoceans} \cite{cmocean-2016}.}

    \label{fig:qual_result}
\end{figure*}
%FIGURE

In general, the output image of a thermal-infrared camera is known to
degrade due to spatial non-uniformities induced from fixed-pattern noise
\cite{ssobarzo-2004}. To solve this problem, most thermal cameras provide
\ac{NUC} using a mechanical shutter. However, every time the \ac{NUC} is
executed, the camera delays by a few seconds and new incoming data may be lost.
This blackout causes fatal defects in motion estimation. Our entire experiment
was performed with the \ac{NUC} disabled. The results show that the effects of
fixed-pattern noise are negligible.

%Most of the experiments were performed within a short time of fewer than five
%minutes

The thermal-\ac{LiDAR} sensor system was mounted on a car-like vehicle
(outdoors) or was hand-held (subterranean). For the outdoor experiment, the
vehicle was also equipped with a VRS-GPS providing accurate positions
(\unit{20}{mm} in a fixed state and \unit{1}{m} in a floating state). Therefore,
we leveraged the measured position at fixed and floating state as the ground
truth when computing the estimation error.

\subsection{Outdoor Experiments}

%Table
\input{tab_exp_kaist.tex}
%Table

The outdoor experiments were conducted on a campus environment including parking
lots, roads, surrounding buildings, and trees. Unlike the RGB image acquired
from the visible camera, the thermal camera detects the infrared energy emitted
or reflected from the object, not being affected by the light source and
applicable regardless of day and night as shown in \figref{fig:environment}.

For evaluation, we collected a dataset consisting of 14 sequences. The traversed
distance of the dataset is \unit{8.8}{km} in total, containing data in the daytime
and night-time. In each sequence, we compared the trajectory of the proposed method
against ORB-SLAM using the global position provided by the VRS-GPS. Since the ground
truth was provided at intervals of  \unit{1}{Hz} and the accuracy is sparsely
guaranteed only in the fixed and float states, we interpolated and aligned the
estimated trajectories based on the timestamp of the VRS-GPS and then calculated
the absolute position error with the global position. In addition, since
ORB-SLAM utilizes only a thermal-camera, an accurate scale cloud not be obtained.
For a fair evaluation, we calculated the absolute position error after
compensating the scale to match the trajectory provided by the VRS-GPS with the
trajectory of the ORB-SLAM. For ORB-SLAM requiring an 8-bit intensity image in
this evaluation, we used an image that rescaled the raw temperature value from
$0^\circ$ to $30^\circ$.

%FIGURE
\begin{figure*}[!t]%
  \centering%
    \includegraphics[width=0.75\textwidth]{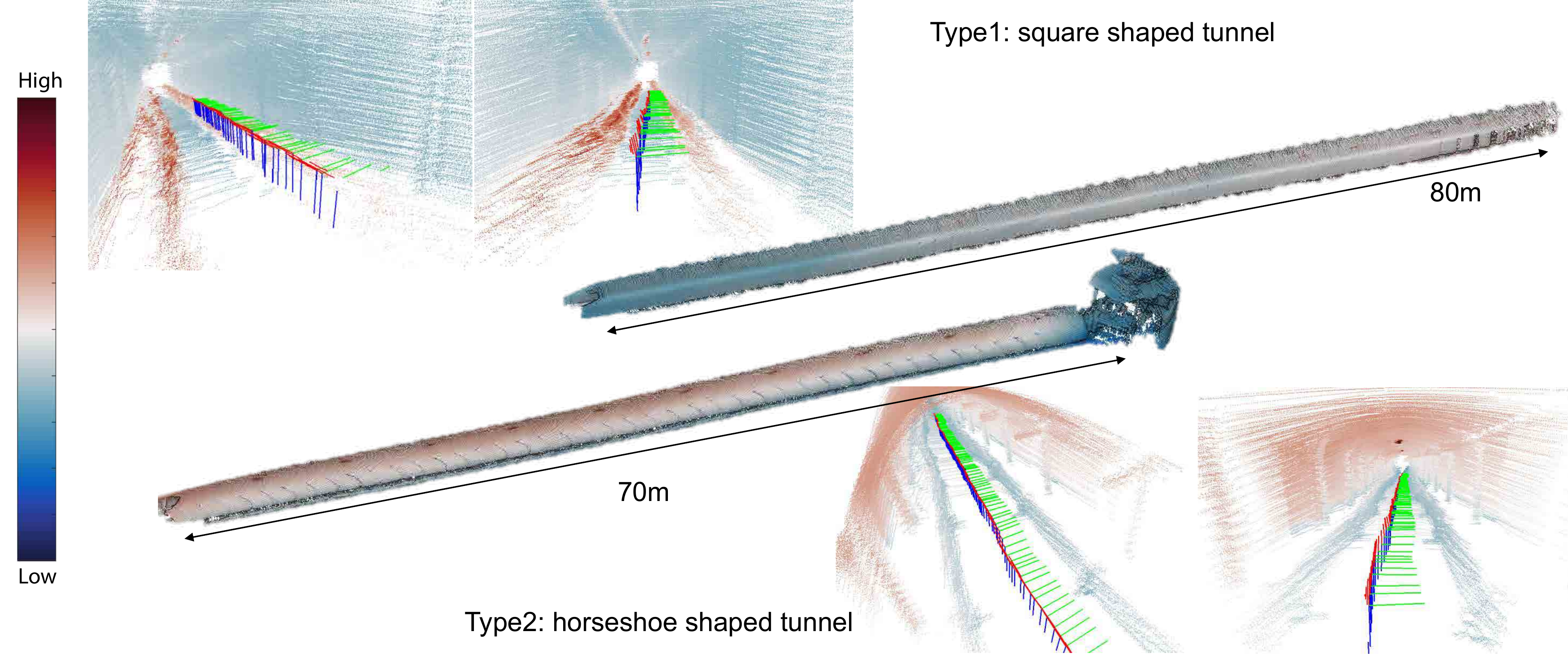}%

  \caption{3D temperature map generated from subterranean tunnels. We performed
  experiments in two tunnels, a horseshoe type tunnel, and a square type tunnel.
  The width and height of the two tunnels are 2.5m. The path lengths are 70m and
  80m, respectively. Our experiments were carried out by moving the
  thermal-LiDAR sensor hand-held and walking in the forward direction. The
  temperature distribution of the two tunnels was quite different. The square
  type tunnel has colder ceiling while the horseshoe type tunnel overall warmer
  temperature as can be seen in the local view of each tunnel. The estimated
  camera coordinates are plotted colored coordinate in the local views. The
  color code is red for the x-axis, green for the y-axis, and blue for the
  z-axis.}

  \label{fig:subterranean_res}%
\end{figure*}
%FIGURE

%All experiments were performed in a commercial laptop environment with Intel Core i7-7700HQ CPU and 8GB RAM.

\figref{fig:qual_result} depicts trajectory estimation and 3D thermal mapping
results in three sequences. The plot was generated excluding GPS-denied regions
where VRS-GPS was unavailable to provide a precise global position at the
cm-level. Our method is not only globally consistent but also provides a smooth
trajectory over all regions. We depicted temperature associated point could
simultaneously performing both 3D thermal mapping and the motion estimation.

\tabref{tl:ape_result} shows the absolute position error between the estimated
position and ground truth. The proposed method outperformed the ORB-SLAM for the
entire sequence. Compared to ORB-SLAM failing at night-time, the proposed method
presents consistent performance both in the daytime and night-time. In both
methods, the absolute position error tends to increase as the traveled distance
increases. This is considered to be due to an accumulation of errors caused by
an increase in traveled distance. For the night time missions, Seq \texttt{04}
to \texttt{06}, rescaling to 8-bit with a fixed range resulted in a low contrast
image and initialization failure for ORB-SLAM. These low contrast images also
induced some larger errors in the ORB-SLAM under rotational motion during Seq
\texttt{03} and \texttt{11}.

%--------------------------------------------------------------------------%
\subsection{Subterranean Experiments}

Secondly, we conducted experiments in two man-made subterranean tunnels, a
horseshoe and a square type tunnel. By hand-holding the sensor systems, the data
was captured by human working along the tunnel. In particular, the square type
tunnel was surveyed in a completely dark environment. Since there is no
available ground-truth, we only show the generated 3D temperature map as a
qualitative result in \figref{fig:subterranean_res}.

%% file: fig_img_samples.tex
\begin{figure*}[!t]
  \centering
  \def\width{0.16\textwidth}%
  \subfigure{%
    \includegraphics[width=\width]{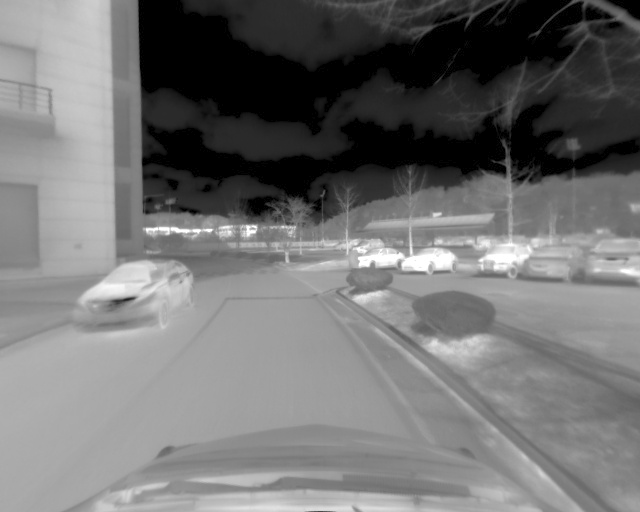}%
    \label{F:thermal_thumbnail1}%
  }
  \subfigure{%
    \includegraphics[width=\width]{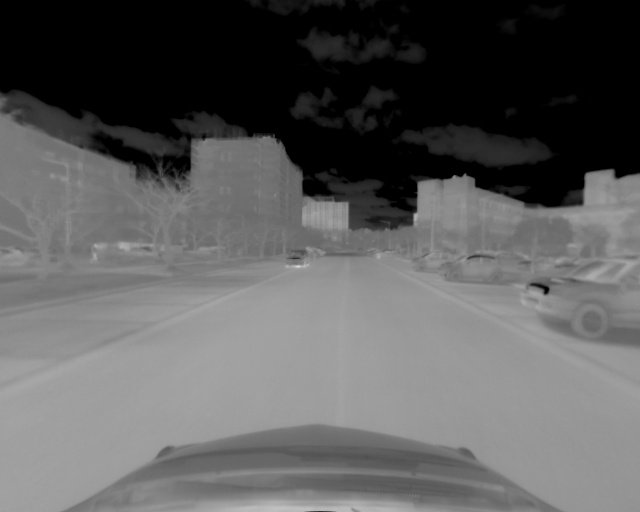}%
    \label{F:thermal_thumbnail2}%
  }
  \subfigure{%
    \includegraphics[width=\width]{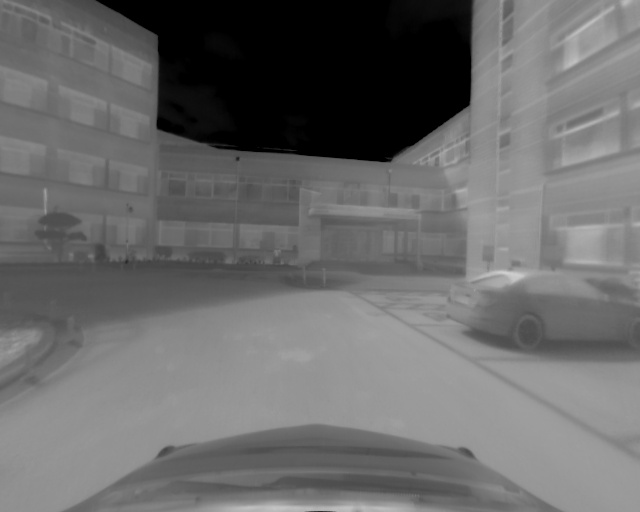}%
    \label{F:thermal_thumbnail3}%
  }
  \subfigure{%
    \includegraphics[width=\width]{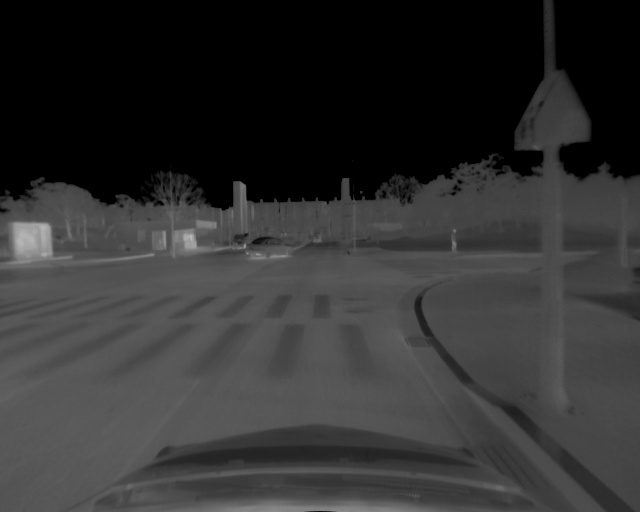}%
    \label{F:thermal_thumbnail4}%
  }
  \subfigure{%
    \includegraphics[width=\width]{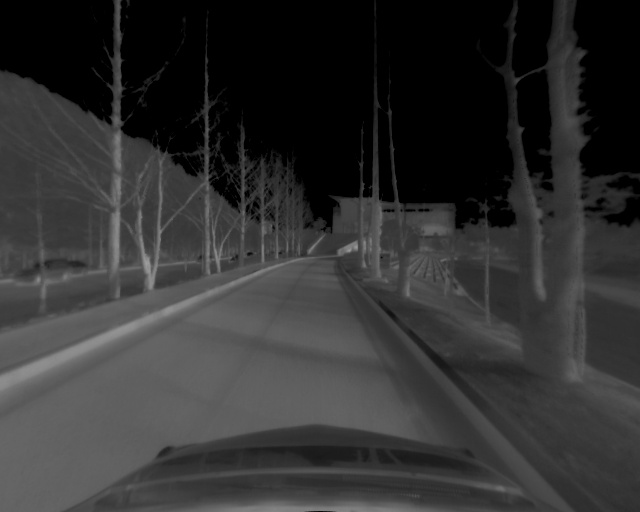}%
    \label{F:thermal_thumbnail5}%
  }\\
  \subfigure{%
    \includegraphics[width=\width]{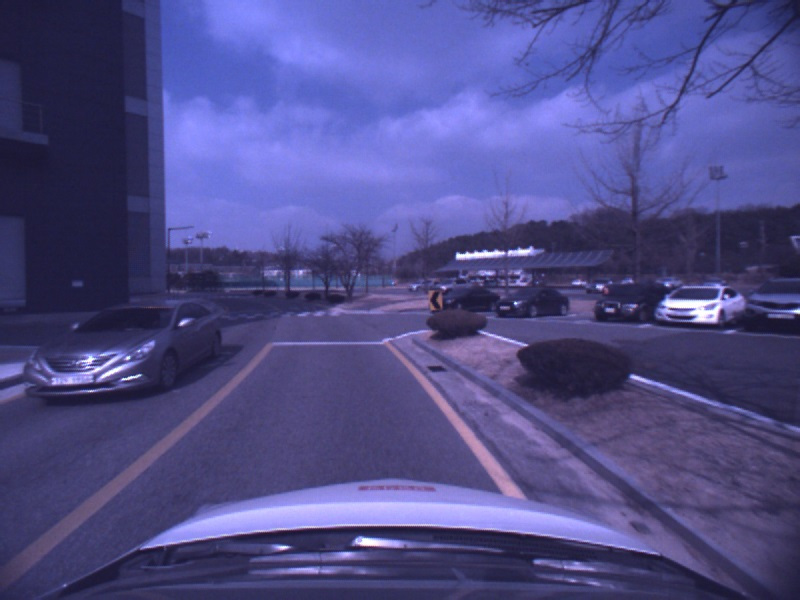}%
    \label{F:rgb_thumbnail1}%
  }
  \subfigure{%
    \includegraphics[width=\width]{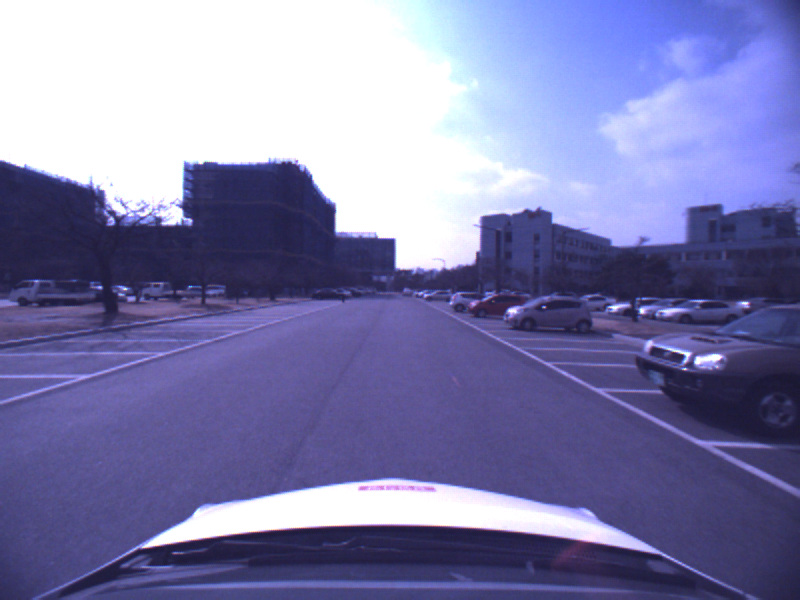}%
    \label{F:rgb_thumbnail2}%
  }
  \subfigure{%
    \includegraphics[width=\width]{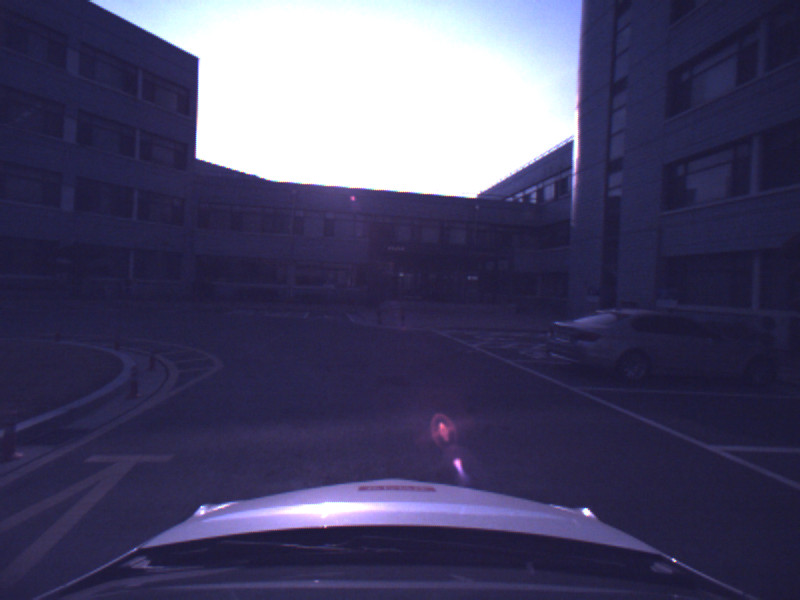}%
    \label{F:rgb_thumbnail3}%
  }
  \subfigure{%
    \includegraphics[width=\width]{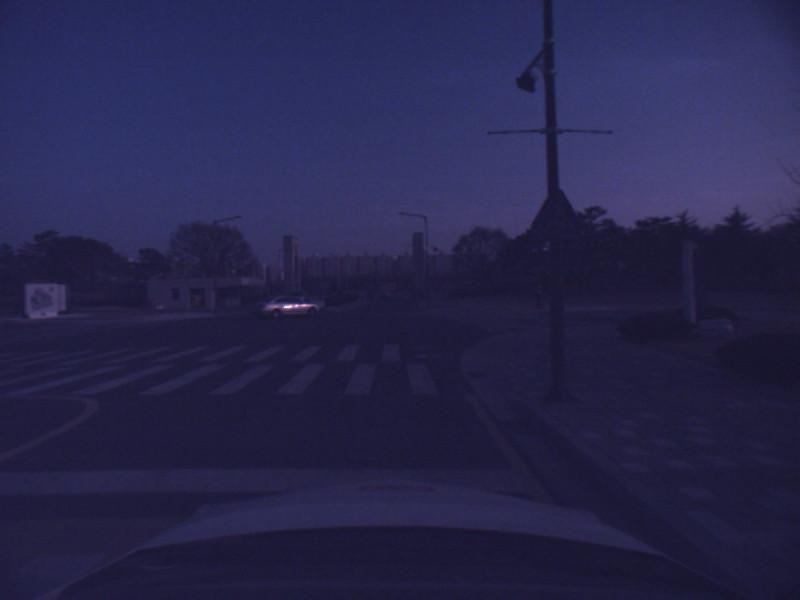}%
    \label{F:rgb_thumbnail4}%
  }
  \subfigure{%
    \includegraphics[width=\width]{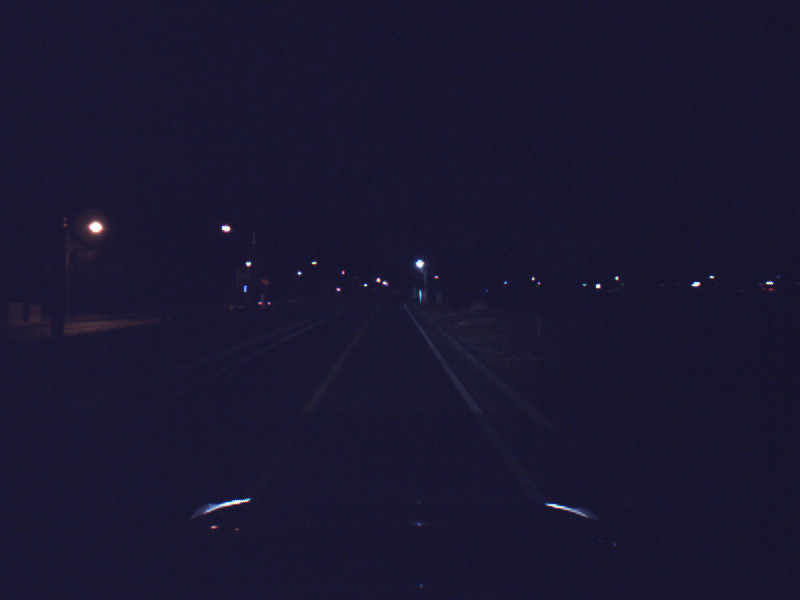}%
    \label{F:rgb_thumbnail5}%
  }

  \caption{Sample thermal-infrared images (first row) and RGB images (second
  row). Unlike RGB images, thermal images are clearly captured regardless of
  light condition (day or night).}

  \label{fig:environment}
\end{figure*}

%% file: tab_exp_kaist.tex
\begin{table*}[]
\centering

\caption{\ac{RMSE} between the VRS-GPS and estimated positions. The blank (-) indicates failures.}

\begin{tabular}{@{}lccrr@{}}
\toprule
\multicolumn{1}{c|}{Seq. Number}            & Traveled distance [m] & Time of day       & \multicolumn{1}{c}{Proposed [m]} & \multicolumn{1}{c}{ORB-SLAM [m]} \\ \midrule\midrule
\multicolumn{1}{l|}{Seq \texttt{01} (2019-01-17-16-27-31)}  & 108   & day (4pm)     & 0.698                       & 1.160                       \\
\multicolumn{1}{l|}{Seq \texttt{02} (2019-01-17-16-47-37)}  & 996   & day (4pm)     & 4.758                       & 6.693                       \\
\multicolumn{1}{l|}{Seq \texttt{03} (2019-01-17-17-15-56)}  & 1031  & day (5pm)     & 5.340                       & 15.553                      \\ \midrule
\multicolumn{1}{l|}{Seq \texttt{04} (2019-01-30-23-28-13)}  & 129   & night (11pm) & 0.690                       & -                            \\
\multicolumn{1}{l|}{Seq \texttt{05} (2019-01-30-23-30-43)}  & 284   & night (11pm) & 1.076                       & -                            \\
\multicolumn{1}{l|}{Seq \texttt{06} (2019-01-31-01-09-33)}  & 835   & night (1am)  & 3.586                       & -                            \\ \midrule
\multicolumn{1}{l|}{Seq \texttt{07} (2019-02-16-13-08-52)}  & 390   & day (1pm)     & 1.521                       & 1.667                       \\
\multicolumn{1}{l|}{Seq \texttt{08} (2019-02-16-13-15-27)}  & 383   & day (1pm)     & 1.264                       & 6.821                       \\
\multicolumn{1}{l|}{Seq \texttt{09} (2019-02-16-16-41-04)}  & 924   & day (4pm)     & 2.931                       & 6.601                       \\
\multicolumn{1}{l|}{Seq \texttt{10} (2019-02-16-16-54-30)}  & 883   & day (4pm)     & 3.519                       & 42.014                      \\
\multicolumn{1}{l|}{Seq \texttt{11} (2019-02-16-17-15-36)}  & 604   & day (5pm)     & 1.818                       & 3.108                       \\ \midrule
\multicolumn{1}{l|}{Seq \texttt{12} (2019-02-16-19-53-31)}  & 365   & night (7pm)  & 1.195                       & -                            \\
\multicolumn{1}{l|}{Seq \texttt{13} (2019-02-16-20-02-51)}  & 1003  & night (8pm)  & 6.398                       & 8.830                       \\
\multicolumn{1}{l|}{Seq \texttt{14} (2019-02-16-20-16-59)}  & 877   & night (8pm)  & 4.193                       & 7.744                       \\ \bottomrule
\end{tabular}
\label{tl:ape_result}
\end{table*}

%% file: conclusion.tex
\section{Conclusions}
\label{sec:conclusion}

We present a direct thermal-infrared \ac{SLAM} that utilizes sparse depth from
\ac{LiDAR}. Our approach is to perform motion estimation by directly tracking
the sparse depth on a 14-bit raw thermal image. Keyframe-based local refinement
improved local accuracy, while loop closure modules enhanced global consistency.
The experiment was conducted with a portable thermal-infrared camera and LiDAR.
Including daytime and night-time and complete darkness, we achieved an all-day
visual SLAM with an accurate trajectory and a 3D thermographic map.

%A natural next step would be 3D temperature map refinement through techniques
%such as sparse pose adjustment \cite{kkonolige-2010}. In addition, we plan to
%handle moving objects to ensure robustness in a dynamic environment.